# BactoBot: A Low-Cost, Bacteria-Inspired Soft Underwater Robot for Marine Exploration

**Authors and Affiliations**

Rubaiyat Tasnim Chowdhury[1]
Nayan Bala[1]
Ronojoy Roy[1]
Tarek Mahmud[1]

[1]Department of Mechanical Engineering, Bangladesh University of Engineering & Technology (BUET), Dhaka, Bangladesh

**Abstract:** Traditional rigid underwater vehicles pose risks to delicate marine ecosystems due to high-speed propellers and rigid hulls. This paper presents BactoBot, a low-cost, soft underwater robot designed for safe and gentle marine exploration. Inspired by the efficient flagellar propulsion of bacteria, BactoBot features 12 flexible, silicone-based arms arranged on a dodecahedral frame. Unlike high-cost research platforms, this prototype was fabricated using accessible DIY methods, including food-grade silicone molding, FDM 3D printing, and off-the-shelf DC motors. A novel multi-stage waterproofing protocol was developed to seal rotating shafts using a grease-filled chamber system, ensuring reliability at low cost. The robot was successfully tested in a controlled aquatic environment, demonstrating stable forward propulsion and turning maneuvers. With a total fabrication cost of approximately $355 USD, this project validates the feasibility of democratizing soft robotics for marine science in resource-constrained settings.

# 1 Introduction

The world's oceans represent vast ecosystems that remain largely unexplored, holding critical data for climate science and biodiversity. While conventional Remotely Operated Vehicles (ROVs) have facilitated significant discoveries, their reliance on rigid hulls and high-speed rotary propellers presents a fundamental limitation: they pose physical risks to sensitive environments such as coral reefs, underwater caves, and algal blooms. To address this, soft robotics has emerged as a transformative paradigm, proposing machines with compliant bodies capable of safe, gentle interaction with natural surroundings. This paper presents **BactoBot**, a low-cost, bio-inspired soft robot designed to bridge the gap between advanced biological mimicry and accessible marine exploration tools.

## 1.1 Background

Research into aquatic robotics has long drawn inspiration from nature to achieve efficient locomotion. Foundational work by **Georgiades et al. (2004)** on amphibious walking robots established early baselines for underwater mobility, while researchers like **Kim et al. (2013)** explored lift-based propulsion inspired by marine turtles. However, these early iterations often retained rigid structures that limited their safe deployment in fragile ecosystems.

Consequently, the field has shifted toward soft materials that replicate natural compliance. **Calisti et al. (2017)** established the fundamental principles of soft locomotion, categorizing the distinct physical paradigms required for soft robots to crawl, swim, and jump effectively. Building on this framework, comprehensive reviews by **Youssef et al. (2022)** and **Qu et al. (2024)** document the rapid evolution of soft robots inspired by macro-scale animals, such as fish, jellyfish, and octopuses. **Qu et al. (2024)** emphasize that soft systems offer superior environmental adaptability compared to their rigid counterparts. Furthermore, **Zhang (2024)** highlights recent advances in nanocomposites that allow robotic fish to mimic muscle-like undulations for stealthy propulsion. Beyond swimming, complex multi-modal concepts have been proposed, such as the amphibious Lizard-Spider-Octopus-Jellyfish robot (LSOJRR) by **Fill et al. (2021)** and the benthic walking robot SILVER2 by **Picardi et al. (2020)**. While effective, these designs often entail high fabrication complexity and reliance on expensive instrumentation, creating barriers to widespread adoption.

A significant gap remains in the literature regarding microorganisms. Despite flagellated bacteria being cited as the "most efficient machines in the universe" for low-Reynolds-number navigation **(Du et al., 2021)**, they are rarely used as prototypes for macro-scale robots. Early attempts to replicate this locomotion were purely mechanical; **Lin (2025)** demonstrated that bacterial swimming patterns could be generated using simple elastic forces. Progressing to electromechanical systems, **Lim et al. (2022)** utilized a centimeter-scale robotic platform coupled with a computational framework to analyze the fluid-structure interactions of flagella. Their work was pivotal in demonstrating how the bundling of soft helical filaments enhances propulsion efficiency and provides robustness against buckling compared to single-flagellated counterparts. More recently, **Hao et al. (2023)** expanded on these physics-based tools to analyze tumbling behaviors for attitude adjustment, though their work focused on fixed-point dynamics rather than free-swimming exploration.

Building on these foundations, recent studies have attempted to integrate advanced control strategies. **Nurzaman et al. (2012)** and **Armanini et al. (2022)** made strides in modeling flagellar motion, with the latter demonstrating that tuning filament stiffness enhances velocity. The state-of-the-art is currently represented by 'ZodiAq' **(Mathew et al., 2025)**, an isotropic flagella-inspired drone. However, while ZodiAq validates the hydrodynamic efficiency of this design, it relies on high-cost custom actuators, keeping the technology out of reach for many research groups.

## 1.2 Problem Statement and Research Questions

Despite the technical progress in bio-inspired robotics, high equipment costs remain a primary barrier to marine science, particularly in developing regions **(Chua et al., 2023)**. As noted by **Xu and Dabiri (2022)**, traditional AUVs and sensor arrays are often prohibitively expensive, with individual units costing tens of thousands of dollars. This makes scalable ocean exploration financially intractable for smaller institutions.

While some recent initiatives have attempted to create low-budget aquatic robots **(Mason and Kelly, 2024)**, effective waterproofing and durability remain persistent failure points in low-cost iterations **(Manos and Kavallieratou, 2025)**. There is currently a lack of proper characterization of DIY methods for replicating complex flagellar locomotion, specifically regarding material selection and reliable waterproofing for rotating shafts in real-life environments.

Addressing these research gaps, this study aims to develop a methodology to create an aquatic robot that answers the following questions:

1. Can a robot be created that replicates bacterial flagellar motion at a very low cost?
2. Is the material used for the flagella (silicone) suitable for real-life aquatic interaction?
3. Can a robust dynamic waterproofing system be developed within a low-cost range?
4. Can the system perform effectively in a real-life water environment rather than only in a controlled lab tank?
5. Can the design be easily replicated by other groups using DIY methods?

### 1.3 Contribution and Real-World Applications

The primary contribution of this paper is the design, fabrication, and validation of **BactoBot**, a soft underwater robot constructed for approximately **$355 USD**. Unlike previous studies that focus on hydrodynamic optimization using expensive hardware, this work demonstrates the feasibility of democratizing soft robotics. We present a novel **multi-stage waterproofing protocol** using a grease-filled chamber system that effectively eliminates the need for expensive mechanical seals.

The real-world applications of these results are significant for environmental monitoring in resource-constrained settings. A low-cost, soft-bodied robot is uniquely suited for the **gentle exploration of fragile ecosystems**, such as monitoring coral bleaching or inspecting underwater infrastructure where bladed propellers pose an entanglement risk. Furthermore, the open-source nature of the design paves the way for educational institutions in developing nations to conduct marine science without requiring grant-level funding.

# 2 Methodology

The development of BactoBot followed a systemic design approach, prioritizing cost-efficiency and reproducibility without compromising the structural integrity required for underwater operation. The system architecture is divided into three primary subsystems: bio-mimetic mechanical structure, electronic control topology, and a novel multi-stage waterproofing protocol.

### 2.1 Bio-Mimetic Mechanical Design

The robot's chassis utilizes a regular dodecahedral geometry, selected to facilitate the symmetrical placement of 12 propulsive arms. This configuration provides inherent kinematic redundancy; in the event of actuator failure, the symmetry of the remaining arms allows for continued, albeit degraded, locomotion. The central frame was fabricated using Fused Deposition Modeling (FDM) 3D printing. Polyethylene Terephthalate Glycol (PETG) was selected as the filament material due to its superior layer adhesion and hydrophobicity compared to Polylactic Acid (PLA), and its higher ductility compared to Acrylonitrile Butadiene Styrene (ABS), which proved too brittle for the internal motor couplings during initial stress testing.

The propulsive flagella were fabricated via a mold-casting process using food-grade silicone rubber to ensure environmental safety. To couple the soft silicone arm with the rigid DC motor shaft, a custom 3D-printed "hook" was embedded into the base of the arm during the curing process. This design explicitly mimics the morphology of the bacterial flagellum, where a rotary nanomachine drives a filament through a universal joint. While biological motors in species like *E. coli* and *Vibrio alginolyticus* reach rotational speeds between 300 and 1700 rpm **(Nakamura and Minamino, 2019)**, BactoBot replicates this functionality at the macroscale. When the initially straight silicone arms are rotated in a fluid medium, hydrodynamic drag forces cause them to deform into helical shapes passively, generating forward thrust through non-reciprocal motion.

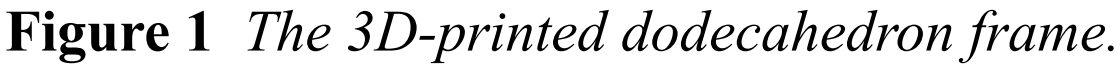

**Figure 1** *The 3D-printed dodecahedron frame.*

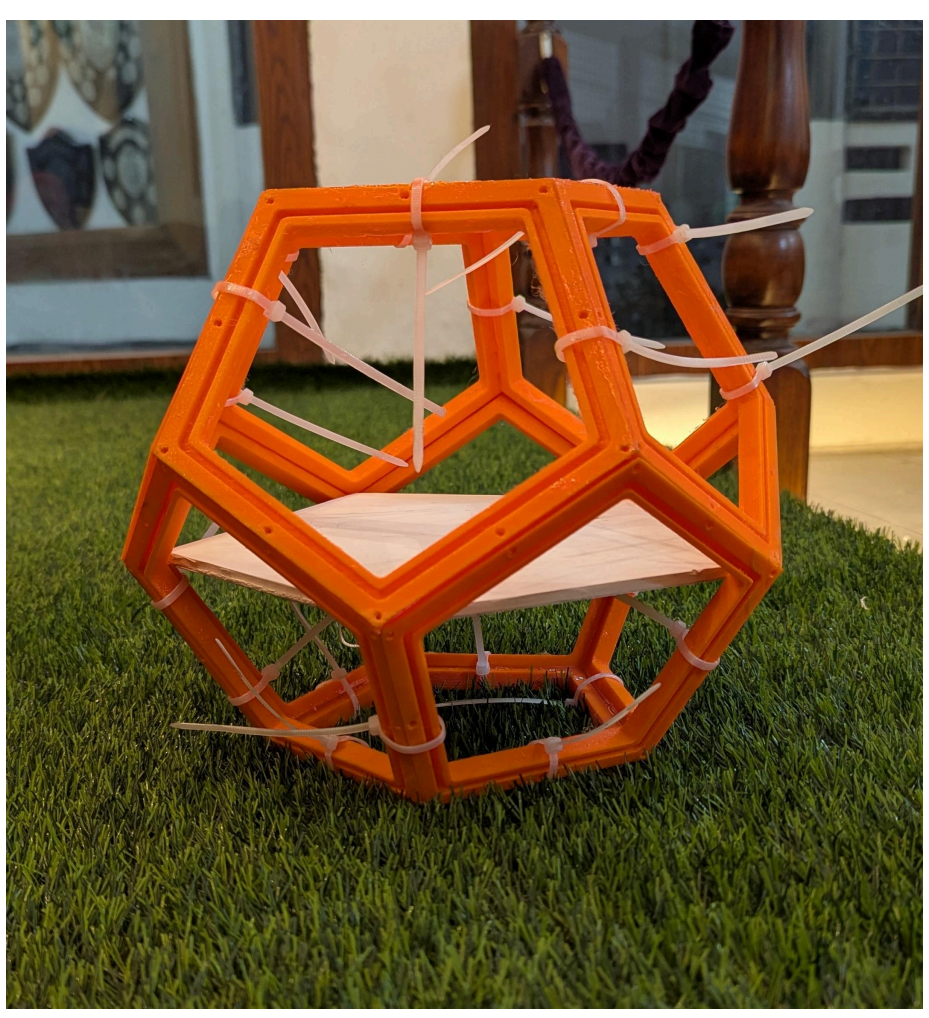

**Figure 2** Design and fabrication of the propulsive flagellum. (a) Technical schematic illustrating the tapered filament geometry (length: 300 mm) and the 45° internal hook angle required for helical deformation. (b) The physical prototype curing within the 3D-printed PETG mold, demonstrating the precise tapered shape achieved through the casting process.

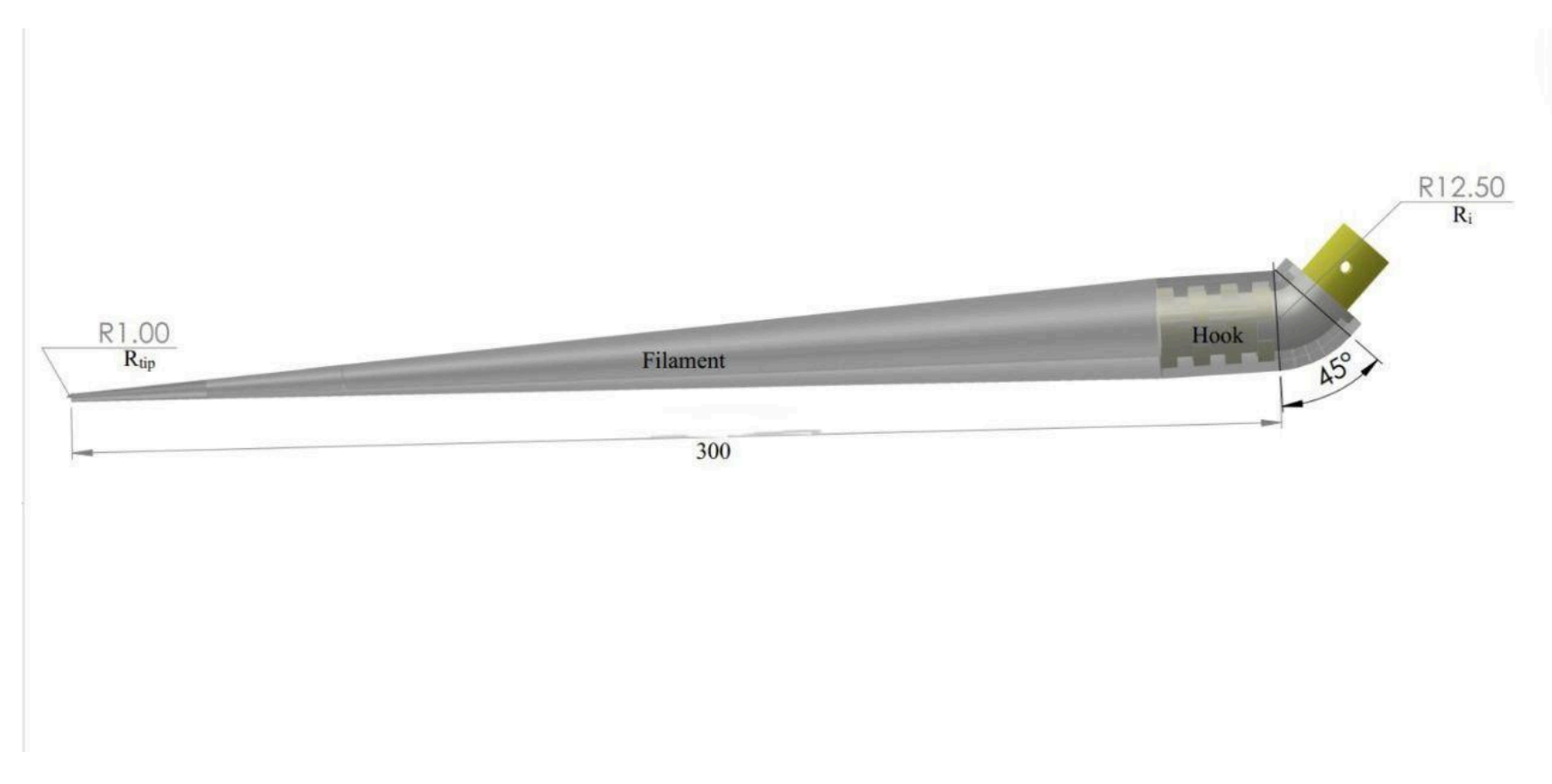

**(a)**

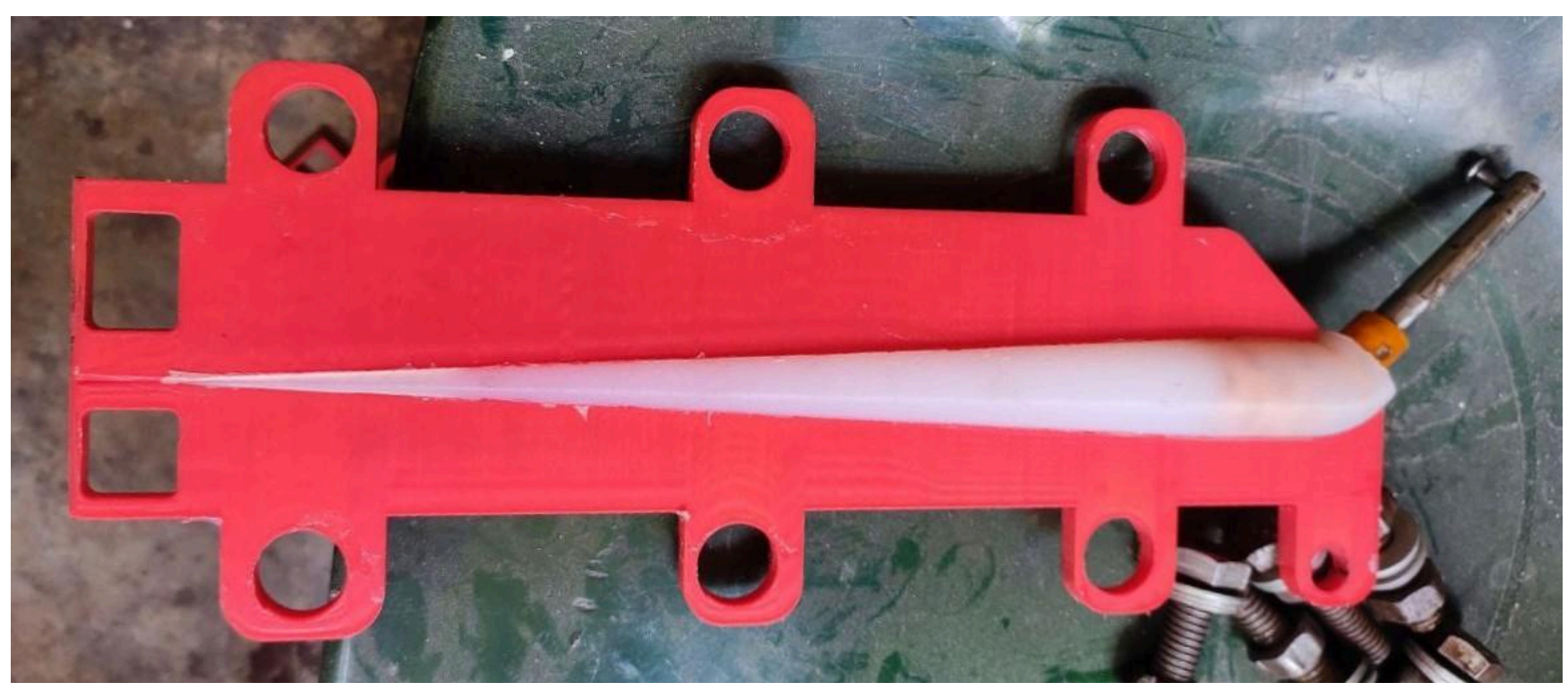

**(b)**

## 2.2 Electronic Architecture and Actuation

The control system is built upon an Arduino Mega 2560 microcontroller, chosen for its extensive Digital I/O capacity. Propulsion is provided by 12 high-torque 12V DC geared motors (100 RPM, 25GA), arranged in six opposing pairs to balance thrust vectors.

Due to the volumetric constraints of the dodecahedral chassis, implementing individual driver circuits for all 12 motors was not spatially feasible. Consequently, a parallel actuation topology was adopted. Each opposing pair of motors is driven by a single BTS7960 High-Current (43A) H-Bridge driver. This driver was selected over the compact TB6612FNG modules used in earlier iterations, which suffered from thermal failure under load. The BTS7960 provides necessary current overhead and thermal protection, ensuring system stability during prolonged operation. Power is supplied by a high-discharge 3S LiPo battery (11.1V, 3300mAh), with voltage regulation for logic circuits managed by a dedicated buck converter.

**Figure 3** *Schematic for a single motor.*

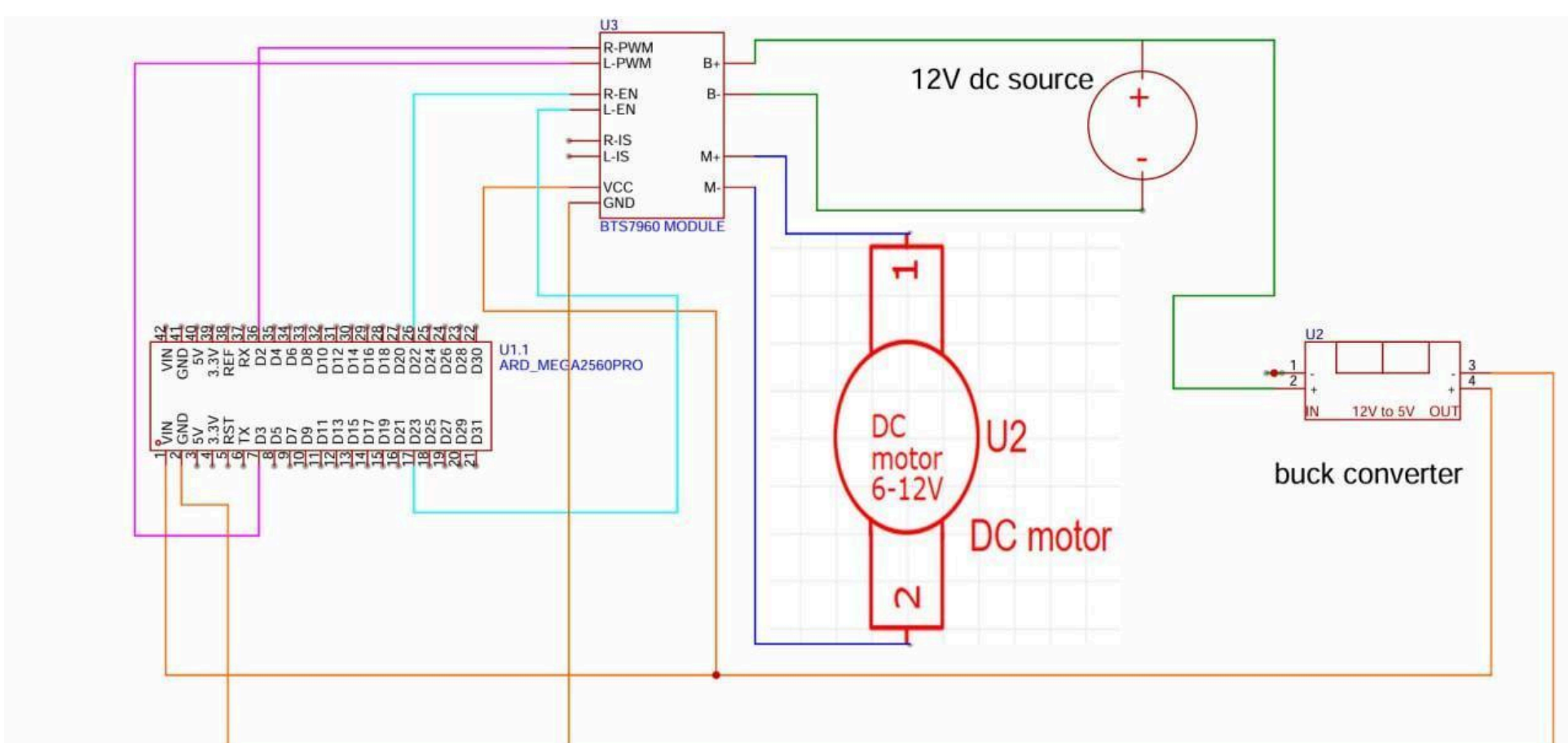

## 2.3 Novel Multi-Stage Waterproofing Protocol

A primary engineering challenge in low-cost underwater robotics is sealing rotating shafts without utilizing expensive ceramic mechanical seals. To address this, we developed a novel, four-stage "Grease-Chamber" waterproofing protocol to ensure dynamic sealing reliability:

1. **Terminal Sealing (Static):** Critical wire ingress points and motor terminals were encapsulated using a two-stage application of Araldite epoxy followed by hot glue. This creates a hermetic seal while providing strain relief against high-frequency vibrations.
2. **Structural Sealing (Static):** All mating seams of the 3D-printed chassis were bonded with marine-grade silicone sealant. To address the porosity inherent in FDM prints, external surfaces were wrapped in a high-density polyethylene layer reinforced with industrial adhesive tape, preventing micro-leakage through print layer lines.
3. **Grease-Chamber Shaft Seal (Dynamic):** The interface between the rotating motor shaft and the external water column is housed within a custom 3D-printed canister containing a multi-O-ring assembly. The void between the O-rings is packed with high-viscosity lithium grease. This creates a hydrostatic barrier that repels water ingress while simultaneously lubricating the shaft, effectively functioning as a low-cost alternative to commercial mechanical seals.
4. **Internal Moisture Control:** To mitigate condensation caused by temperature differentials, oil-impregnated open-cell foam was installed within the main electronics housing. This serves to trap residual atmospheric moisture and provides passive lubrication to internal components.

## 2.4 Buoyancy Calibration

The robot utilizes a neutral buoyancy strategy to minimize energy consumption during station-keeping. The calibration process involved a three-step iterative approach:

1. **Digital Estimation:** The displacement volume was calculated using SolidWorks CAD analysis to establish a baseline.

2. **Mass Prediction:** Theoretical computation was cross-referenced with material density data to estimate the required ballast.
3. **Iterative Tank Testing:** The prototype was submerged in a test tank, and lead ballast weights were added in 500g and 1kg increments until equilibrium was reached.

The system achieved stable neutral buoyancy at a total mass of **11.25 kg**. To ensure stability, the center of mass was mechanically lowered by positioning the battery and heaviest ballast at the bottom of the chassis. This provides a passive righting moment (metacentric stability), preventing uncontrolled rolling during operation.

# 3 Results and Analysis

## 3.1 Cost Analysis and Economic Viability

A primary objective of this research was to validate that soft underwater robotics can be democratized through cost reduction. **Table 1** presents the cost breakdown by subsystem. The total fabrication cost of the BactoBot prototype was approximately **43,359 BDT** (approx. **$355 USD**).

**Table 1** Summary of overall fabrication costs

| **Category** | **Cost(BDT)** | **Cost(USD)*** | **% of Total** |
|---|---|---|---|
| Structural Mechanics (Silicone, 3D Printing, Frame) | 23,140.00 | 189.00 | 53.40% |
| Electrical System (Microcontroller, Drivers, Motors) | 16,568.00 | 135.00 | 38.20% |
| Waterproofing (Sealants, O-Rings, Grease) | 2,331.00 | 19.00 | 5.40% |
| Buoyancy Control (Ballast Weights) | 400.00 | 3.50 | 0.90% |
| Miscellaneous (Consumables, Transport) | 920.00 | 7.50 | 2.10% |
| **Total** | **43,359.00** | **~355** | **100%** |

**Exchange rate approx. 1 USD = 122.60 BDT (as of November 2025).*

The structural components constituted the largest expense (53.4%), driven by the cost of food-grade silicone rubber and the significant volume of PETG filament required for the dodecahedral chassis. Notably, the waterproofing system—typically the most capital-intensive aspect of underwater engineering due to precision machining requirements—accounted for only 5.4% of the budget ($20 USD). This validated the economic efficiency of the novel grease-chamber design.

**Economic Comparison:** To contextualize these figures, the cost of BactoBot must be compared against existing oceanographic tools. As reported by **Xu and Dabiri (2022)**, operational costs for traditional propeller-driven AUVs often exceed **$50,000 USD.** Even within the "low−cost" market segment, **Chua et al.(2023)** identify the open−source *BlueROV2* as the standard, with a price point of approximately **$5,600 USD**. Consequently, the methodology presented here represents a **93% cost reduction** compared to entry-level commercial standards. This confirms that by substituting specialized marine hardware with industrial off-the-shelf alternatives, bio-inspired robotics can be made financially viable for resource-constrained research groups.

## 3.2 Locomotion Performance

Field testing was conducted in a freshwater tank to validate the bio-inspired propulsive mechanism. Upon activation of the DC motors, the flexible silicone arms successfully engaged with the fluid medium. As predicted by the design simulations, hydrodynamic drag caused the initially straight appendages to passively deform into stable helical shapes. This deformation facilitated the generation of thrust through non-reciprocal motion, qualitatively validating the replication of bacterial flagellar mechanics using soft materials.

1. **Forward Translation:** Linear motion was achieved by activating opposing pairs of motors with symmetrical drive signals. The robot demonstrated consistent forward translation with minimal turbulence. Unlike the cavitation and shear forces associated with rigid propellers, the soft flagellar motion produced a gentle wake, a critical requirement for operating near sensitive benthic life.
2. **Stability and Maneuverability:** Throughout the trials, the robot maintained a stable orientation with no uncontrolled rolling or

pitching. This confirms the efficacy of the center-of-mass adjustments made during the buoyancy calibration phase. Directional control (yaw) was successfully demonstrated by generating a net torque around the vertical axis. This was achieved by driving opposing motor pairs in contrary rotational directions (CW vs. CCW), creating a couple-moment that pivoted the robot in place. This validates the control topology's ability to perform the precise heading adjustments required for visual inspection tasks.

**Figure 4** Free-body diagrams illustrating the vector summation of hydrodynamic forces governing BactoBot's locomotion. (a) Forward Translation: Symmetrical actuation of opposing flagellar pairs generates thrust vectors where lateral components cancel out, resulting in a net forward velocity. (b) Directional Turning (Yaw): Differential actuation creates a force couple, generating a resultant torque around the vertical axis to change the robot's heading.

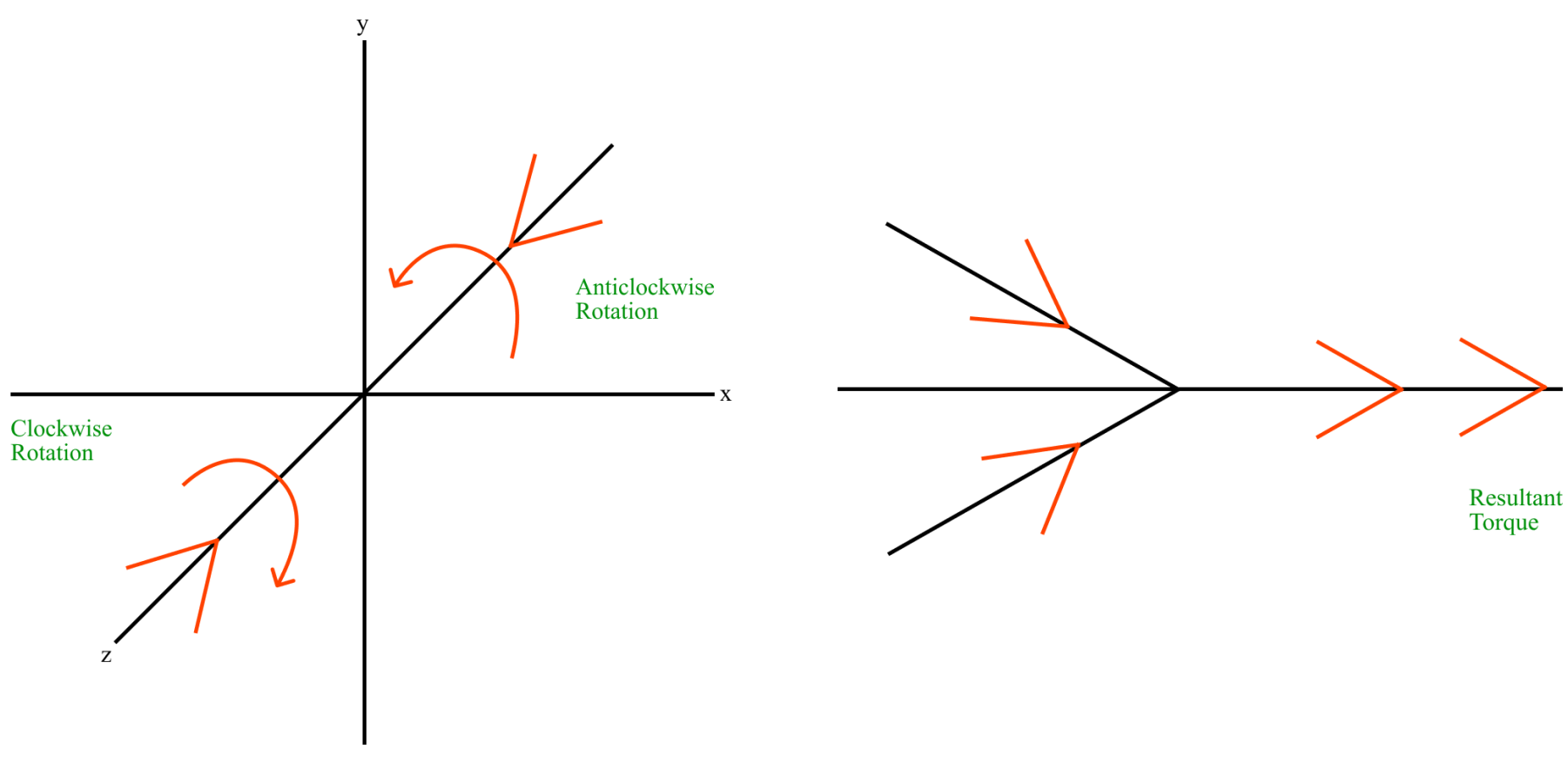

**(a)**

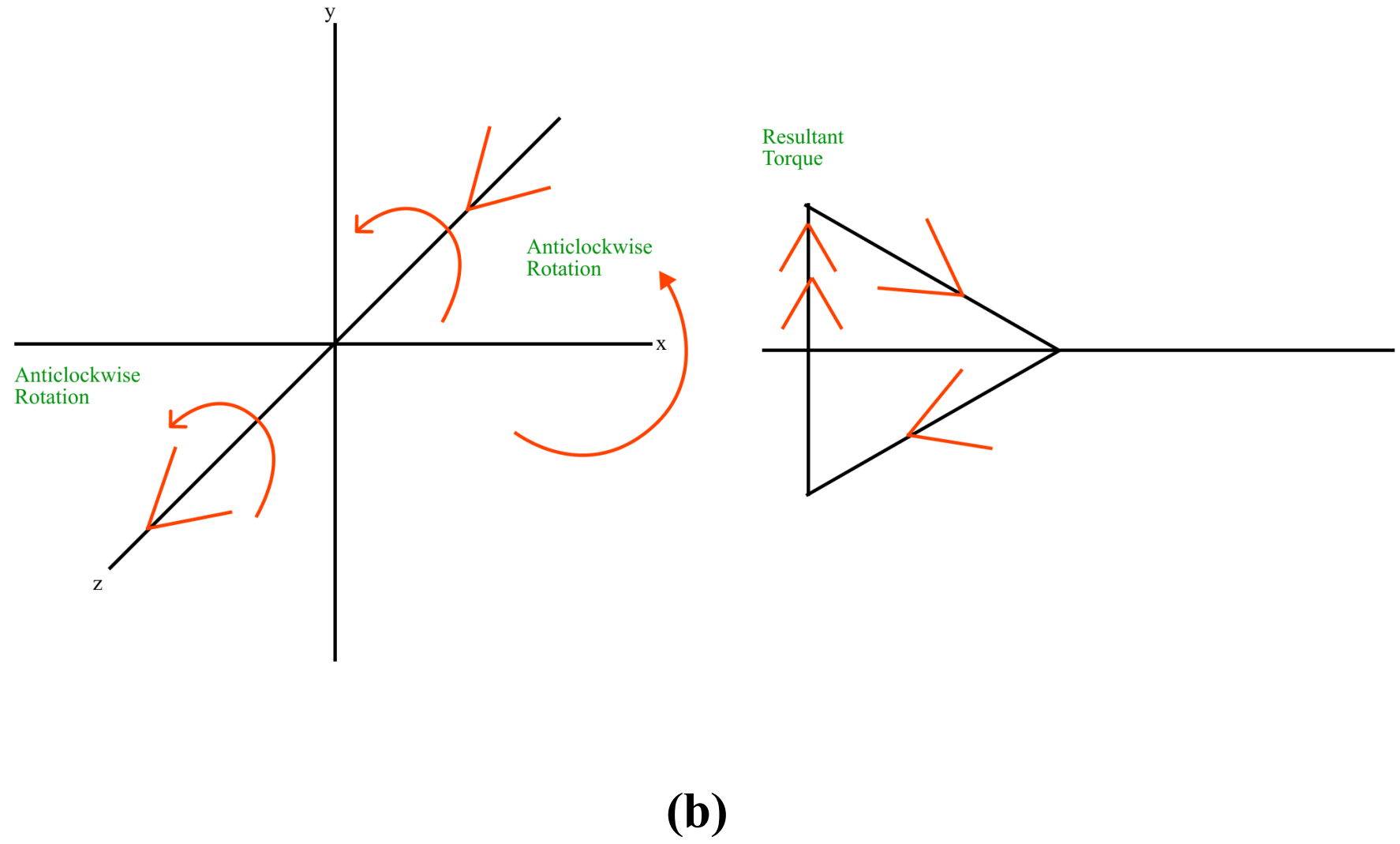

**(b)**

**Figure 5** Photographic results from aquatic field testing. (a) Surface view demonstrating the robot's stable neutral buoyancy and upright orientation after ballast calibration. (b) Submerged operation view, confirming the efficacy of the "Grease-Chamber" waterproofing protocol and the successful actuation of flagella at operational depth.

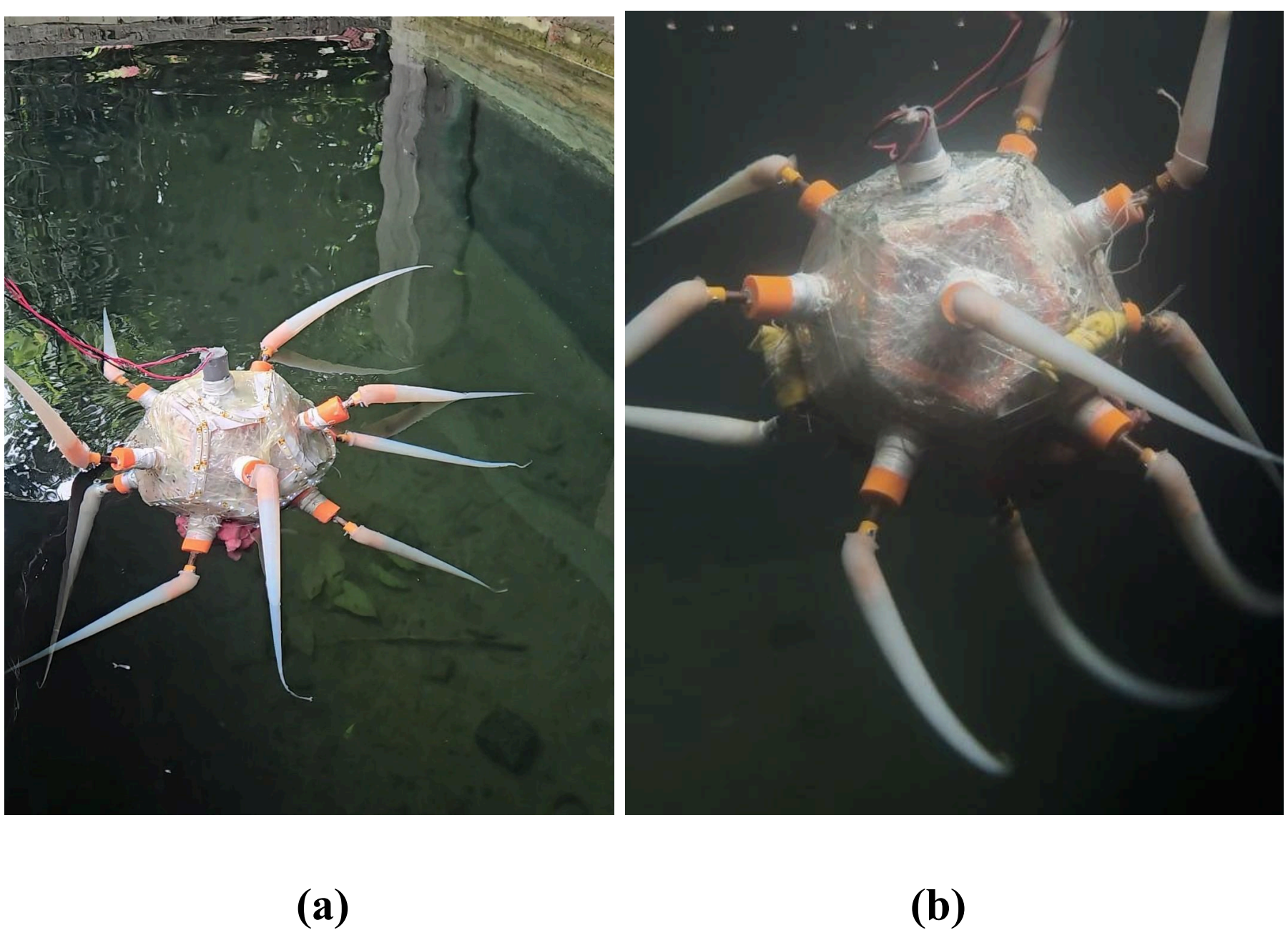

**(a)** **(b)**

### 3.3 Waterproofing Validation

The novel "Grease-Chamber" waterproofing protocol proved highly effective during dynamic operational testing. The robot was subjected to cumulative testing periods exceeding 30 minutes at depths of up to 1 meter. Post-trial inspection revealed zero water ingress within the central electronics housing or the individual motor canisters. This result empirically validates that high-viscosity grease barriers, when housed in properly toleranced 3D-printed canisters, can serve as a reliable, low-cost alternative to ceramic mechanical seals for shallow-water.

# 4 Conclusion

This project successfully met its objectives to design, fabricate, and test a low-cost, bacteria-inspired soft underwater robot. We demonstrated that the complex mechanics of flagellar propulsion can be effectively replicated using accessible DIY methods, specifically FDM 3D printing and silicone molding. **BactoBot** stands as a functional proof-of-concept, validating a fabrication methodology that lowers the cost barrier for marine exploration tools to under **$400 USD**. The successful implementation of the mechanical design, the multi-layered waterproofing protocol, and the buoyancy control system confirms the fundamental viability of this approach for environmental monitoring in sensitive ecosystems where traditional rigid ROVs pose a risk of damage.

### 4.1 Lessons Learned

During the iterative design process, several critical engineering insights were gained that are valuable for future low-cost robotics development:

1. **Material Selection:** The initial use of ABS plastic for the flagellum's internal hook, based on prior literature (**Armanini et al., 2022)**, resulted in mechanical failure due to brittleness. Substituting ABS with PETG provided the necessary ductility to withstand the torque of the motors, highlighting the importance of material properties over precedent in soft-rigid coupling.

2. **Silicone Molding Protocol:** Initial fabrication attempts yielded faulty flagella due to residual moisture in the molds and imprecise mixing ratios. Establishing a strict protocol of ensuring molds were completely dry and meticulously measuring all components by weight was essential for successful and repeatable fabrication.

3. **Electrical Robustness:** To conserve space, compact Dual TB6612FNG motor drivers were initially selected. However, their inadequate current handling led to thermal failure and microcontroller damage. Upgrading to robust BTS7960 H-Bridge drivers resolved these stability issues, demonstrating that electrical safety margins should take precedence over space constraints in underwater systems.

## 4.2 Future Work

While the current prototype validates the mechanical and hydrodynamic principles, the control system currently operates in an open-loop configuration. Future iterations of BactoBot will focus on two key areas:

1. **Closed-Loop Control:** We intend to integrate an Inertial Measurement Unit (IMU) to provide real-time feedback on orientation. This will enable the implementation of a **PID control algorithm** to actively correct for drift and maintain heading against currents.
2. **Autonomous Perception:** The integration of a waterproof camera and computer vision algorithms will be explored to enable obstacle avoidance and visual station-keeping, allowing the robot to perform autonomous data collection tasks in complex environments like coral reefs.

## Disclaimer

This work was conducted as a project for the ME366: Electro-Mechanical System Design and Practice course at the Bangladesh University of Engineering & Technology (BUET). No funding was received to support this research.